\pdfoutput=1

\documentclass[11pt]{article}

\usepackage[]{ACL2023}

\usepackage{times}
\usepackage{latexsym}
\usepackage{booktabs}
\usepackage{multirow}
\usepackage{listings}
\usepackage{amssymb}
\usepackage{amsfonts}
\usepackage{graphicx, xcolor}
\usepackage[table,xcdraw]{xcolor}
\usepackage{MnSymbol}
\usepackage{booktabs}
\usepackage{array}    

\usepackage[T1]{fontenc}

\usepackage[utf8]{inputenc}

\usepackage{microtype}

\usepackage{inconsolata}

\usepackage{makecell}
\usepackage{pgfplots}
\pgfplotsset{compat=1.17}

\title{The Privileged Students: On the Value of Initialization in Multilingual Knowledge Distillation}

\author{
Haryo Akbarianto Wibowo, Thamar Solorio, Alham Fikri Aji \\
\textbf{MBZUAI}\\
\texttt{\{haryo.wibowo,thamar.solorio,alham.fikri\}@mbzuai.ac.ae}
}

\begin{document}
\maketitle
\begin{abstract}

Knowledge distillation (KD) has proven to be a successful strategy to improve the performance of smaller models in many NLP tasks. However, most of the work in KD only explores monolingual scenarios. In this paper, we investigate the value of KD in multilingual settings. We find the significance of KD and model initialization by analyzing how well the student model acquires multilingual knowledge from the teacher model. Our proposed method emphasizes copying the teacher model's weights directly to the student model to enhance initialization. Our findings show that model initialization using copy-weight from the fine-tuned teacher contributes the most compared to the distillation process itself across various multilingual settings. Furthermore, we demonstrate that efficient weight initialization preserves multilingual capabilities even in low-resource scenarios.

\end{abstract}
\section{Introduction}

Multilingual models face challenges related to both data availability and computational resources~\cite{ruder2022statemultilingualai,nityasya2021costs,aji-etal-2022-one}, which makes producing lightweight multilingual models challenging. Recent work has proposed utilizing Knowledge Distillation (KD) in multilingual contexts~\cite{ansell-etal-2023-distilling}. However, due to data scarcity, questions arise about how to effectively transfer knowledge between teacher and student models.

The multilingual KD approach follows the training methodology of TinyBERT~\cite{jiao-etal-2020-tinybert}, involves a two-step distillation process where a smaller student model is pre-trained before fine-tuning to acquire multilingual knowledge from a larger teacher model. However, this method requires substantial data for the pre-training step, thus increasing computational costs. We argue that this pre-training step is crucial to make the KD work in this multilingual scenario, as the scenario requires strong cross-lingual prior knowledge as an initialization to perform well, especially in low-resource scenarios~\cite{conneau-etal-2020-unsupervised}. This initialization serves as important inductive biases for downstream tasks across multiple languages. 

To investigate this, we analyze the impact of both KD and model initialization on multilingual model performance. We propose an efficient initialization method that involves alternately copying weights from the teacher model to the student model, similar to the approach used in DistilBERT \cite{sanh2020distilbert}. While weight copying has been employed previously, its effectiveness in multilingual scenarios remains underexplored. Our study aims to comprehensively examine how KD and weight copying affect multilingual model performance, learning speed, and cross-lingual generalization.

\begin{figure*}
    \centering
    \includegraphics[width=1\linewidth]{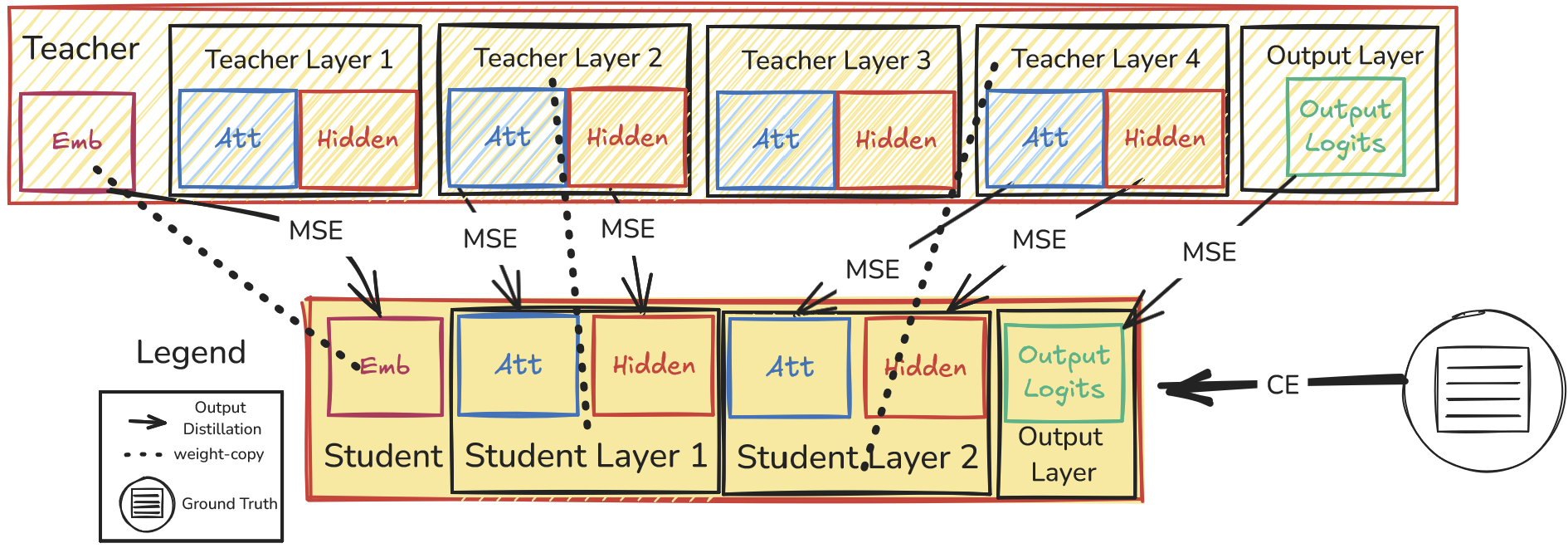}
    \caption{Overall architecture of Knowledge Distillation used in this paper where the teacher distills its knowledge using Mean Square Loss (MSE) and followed by Cross Entropy (CE) Loss with respect to the ground truth from the labeled dataset. This is an example of pair of student-teacher models that have 2 and 4 layers, respectively. Att denotes attention output, Hidden denotes hidden output, and Emb denotes Embedding weights.}
    \label{fig:arch}
\end{figure*}
Our contributions to this work are as follows:

\begin{enumerate}

\item We demonstrate that fine-tuning a well-initialized model, specifically by copying the teacher model's weights, significantly impacts downstream task performance more than leveraging the distillation objective function across various multilingual settings.
\item We show that weight-copy enables zero-shot cross-lingual generalization, preserving multilingual capabilities even in low-resource scenarios.
\item Our experiments reveal that the weight-copy approach leads to faster learning speeds and exhibits multilingual knowledge. Among the weight-copying strategies we compared, directly copying the teacher's weights resulted in the fastest learning speed across different data sizes.
\end{enumerate}

\section{Methodology} \label{sec: methodology}

Knowledge distillation (KD) is a technique used to transfer knowledge from a large, trained teacher model to a smaller student model. This process aims to retain the large model's performance while reducing the computational cost during inference. KD involves training the student model to mimic the outputs of the teacher model, often using a combination of the teacher’s soft target outputs and the ground truth labels.

Various extensions of KD have been proposed to enhance its effectiveness. For instance, TinyBERT~\citep{jiao-etal-2020-tinybert, ansell-etal-2023-distilling} introduces a two-step distillation process. First, the student model is pre-trained on a large corpus to acquire good initialization suitable for the next step. Afterward, the model is fine-tuned for the desired tasks. This approach requires substantial data and computational resources, making it challenging to implement with limited resources.

In this work, we aim to explore the impact of these components of knowledge distillation in multilingual settings, focusing on efficiency. Instead of the extensive pre-training step used in~\citealp{jiao-etal-2020-tinybert, ansell-etal-2023-distilling}, we employ a simpler and more efficient initialization approach by copying the weights from the teacher model to the student model, inspired by DistilBERT~\cite{sanh2020distilbert}. 

\subsection{Distillation Architecture} \label{sec: kd-arch}

We utilize knowledge distillation (KD), comprised of a teacher $T$ and a student $S$ model. The student model always has fewer layers than the teacher model. We follow TinyBERT~\cite{jiao-etal-2020-tinybert}'s objective loss and architecture. The loss of the KD comprises embedding loss $\mathcal{L}_{embd}$, hidden-layer loss $\mathcal{L}_{hidn}$, attention loss $\mathcal{L}_{att}$, and prediction-layer loss $\mathcal{L}_{pred}$. These objective functions can be formulated as follows:

\begin{equation}
    \mathcal{L}_{att} = \frac{1}{l}\frac{1}{h}\sum^l_{i=1}\sum^h_{j=1}\text{MSE}(A_{S}^{i}, A_T^{k})
\end{equation}

\begin{equation}
    \mathcal{L}_{hid} = \frac{1}{l}\sum^l_{i=1}\text{MSE}(W \cdot H_S^{i}, H_T^{k})
\end{equation}

\begin{equation}
    \mathcal{L}_{embd} =\text{MSE}(W \cdot E_{S}, E_{T})
\end{equation}

\begin{equation}
    \mathcal{L}_{pred} =\text{MSE}(z_S, z_T)
\end{equation}

\begin{table*}
\centering
\small
\begin{tabular}{llll}
\toprule
Information & \texttt{massive} & \texttt{tsm} & \texttt{universal-ner} \\
\midrule
Number of Training Data & 11,514 & 1,839 & 6,366* \\
Number of Classes & 60 & 3 & 7 \\
Number of Languages & 52 & 8 & 9 \\
\texttt{unseen lang} partition & 
\begin{tabular}[t]{@{}p{4cm}@{}}
\raggedright "am-ET", "cy-GB", "af-ZA", "km-KH", "sw-KE", "mn-MN", "tl-PH", "kn-IN", "te-IN", "sq-AL", "ur-PK", "az-AZ", "ml-IN", "ms-MY", "ca-ES", "sl-SL", "sv-SE", "ta-IN", "nl-NL", "it-IT", "he-IL", "pl-PL", "da-DK", "nb-NO", "ro-RO", "th-TH", "fa-IR"
\end{tabular} & 
\begin{tabular}[t]{@{}p{3cm}@{}}
\raggedright "arabic", "french", "hindi", "portuguese"
\end{tabular} & 
\begin{tabular}[t]{@{}p{3cm}@{}}
\raggedright "da-ddt", "pt-bosque", "sr-set", "sk-snk", "sv-talbanken"
\end{tabular} \\
\bottomrule
\end{tabular}
\caption{Data statistics for \texttt{massive}, \texttt{tsm}, and \texttt{universal-ner}. Each language consists of the same number of instances in both datasets, except \texttt{universal-ner} which number instance varies across languages. \texttt{unseen lang} denotes language subset used in the zero-shot cross-lingual experiment. The rest of the languages are categorized as \texttt{seen lang}. \texttt{universal-ner} data does not include partition that does not have training set. \textbf{*} denotes the mean of number of instances across languages.}
\label{tab:data-stat}
\end{table*}

Where $A$, $H$, $E$, and $z$ are the values of the attention outputs, hidden layers' outputs, embedding layer's outputs, and the logits, respectively, for the teacher $T$ or student $S$ models. Indices of model layers and attention heads are denoted as $l$ and $h$. If the student model's hidden unit dimension is smaller than the teacher's, we leverage a projection weight $W$ to match the hidden unit dimension. Otherwise, $W$ is an identity matrix\footnote{In the implementation, we omit $W$ instead.}. 

The mapping function of student and teacher model is based on the best ablation results of \citet{jiao-etal-2020-tinybert}, which defined as follows:

\begin{equation}
k = i \cdot \frac{N_T}{N_S}, \quad i \in [1, N_S], \quad k \in [1, N_T]
\end{equation}

In this formula, $i$ represents the index of the student's layer, while $k$ is the index of the corresponding teacher's layer. $N_S$ denotes the total number of layers in the student model, and $N_T$ represents the total number of layers in the teacher model. 

The KD loss can be formulated as follows:

\[
\mathcal{L_{KD}} = \mathcal{L}_{att} + \mathcal{L}_{hid} + \mathcal{L}_{embd} + \mathcal{L}_{pred}
\]

We calculate the classification loss $\mathcal{L}_{clf}$ as follows:

\[
\mathcal{L}_{clf} = CE(z^{S}, GT)
\]

Where $GT$ is the ground truth of the observed instance. 

Finally, we obtain the overall loss $\mathcal{L}_{overall}$ which is going to be minimized in the training process:

\[
    \mathcal{L}_{overall} = \mathcal{L_{KD}} + \mathcal{L}_{clf}
\]

We use Mean Square Error (MSE) instead of KL Divergence due to faster convergence and higher performance, as supported by the experiment of \citealp{nityasya2022student}. The overall architecture can be seen in Figure \ref{fig:arch}.

\subsection{Model Initialization}  \label{sec:model-init}

To avoid the expensive pre-training step used by \citealp{jiao-etal-2020-tinybert, ansell-etal-2023-distilling}, we adopt the model initialization approach from DistilBERT \cite{sanh2020distilbert}, where the student model's weights are initialized by copying the weights of the teacher model.

We alternately copy the weights of the teacher's embedding layer and classification layers to the student model. For the self-attention layer, we copy the weights based on the following mapping function:

\begin{equation}
    \text{SA}_T^j = \text{SA}_S^{i*2} \quad \text{for} \quad i, j \in \mathbb{Z}^+
\end{equation}

Here, SA denotes the self-attention layers of the teacher $T$ and student $S$ models, respectively. The notations $i$ and $j$ indicate the indices of the student and teacher self-attention layers, respectively. To illustrate, the second self-attention layer of the teacher model will be mapped to the first self-attention layer of the student model.

If the student's hidden unit dimension is smaller than the teacher's, we follow the approach of \citealp{xu2024initializing} by selecting evenly spaced elements in the teacher's linear weight and bias for the self-attention layer to map the student's self-attention layer correspondingly. For instance, suppose the teacher has a linear weight of 4x4, and the student has a 2x2 matrix; we select the 1st and 3rd slices along both the first and second dimensions. For the bias, we do the slicing in one dimension instead. 

\begin{table*}
\centering
\small
\setlength{\tabcolsep}{4pt}
\begin{tabular}{@{}llrccccccr@{}}
\toprule
\multirow{2}{*}{\textbf{Model}} & \multirow{2}{*}{\textbf{Method}} & \multirow{2}{*}{\textbf{Layers}} & \multicolumn{3}{c}{\textbf{XLM-R}} & \multicolumn{3}{c}{\textbf{mDeBERTa}} & \multirow{2}{*}{\textbf{Avg}} \\
\cmidrule(lr){4-6} \cmidrule(lr){7-9}
& & & \texttt{massive} & \texttt{tsm} & \texttt{universal-ner} & \texttt{massive} &\texttt{tsm} & \texttt{universal-ner} & \\
\midrule
Teacher & \texttt{from-base} & 12 & 80.18 & 70.10 & 87.66 & 81.36 & 66.96 & 88.81 & 79.18 \\
\midrule
\multirow{4}{*}{Student} & \texttt{from-scratch} & 6 & 75.19 & 50.20 & 45.68 & 75.93 & 49.86 & 46.21 & 57.18 \\
& \texttt{from-scratch} + KD & 6 & 79.23 & 54.13 & 48.29 & 76.22 & 51.80 & 39.55 & 60.30 \\
\cmidrule{2-10}
 & \texttt{from-teacher} & 6 & 81.18 & 62.99 & 79.50 & 80.45 & 58.12 & 78.20 & 73.41 \\
& \texttt{from-teacher} + KD & 6 & \textbf{81.63} & \textbf{67.61} & \textbf{80.18} & \textbf{82.31} & \textbf{61.08} & \textbf{79.17} & \textbf{75.33} \\
\bottomrule
\end{tabular}
\caption{Comparative performance (F1-scores \%) of XLM-R and mDeBERTa models across different datasets (all languages), initialization methods, and with/without knowledge distillation (KD). The teacher model has 12 layers, while all student models have 6 layers. All models are trained with all languages for each data. Bold denotes best methods for each data and their average.}
\label{tab:main-table-result}
\end{table*}

\section{Experiment Setup} \label{sec: exp-setup}

We provide three experiment setups: data, model, and training, that will be consistently used throughout this work. 

\paragraph{Data} We utilized \texttt{massive}~\cite{fitzgerald2022massive}, Tweet Sentiment Multilingual dataset (denoted as \texttt{tsm})~\cite{barbieri-etal-2022-xlm}, and \texttt{universal-ner}~\cite{mayhew2024universal}. We selected these datasets to observe the behavior of multilingual performance under different situations: high-resource data with parallel data (\texttt{massive}) and low-resource data with non-parallel data (\texttt{tsm}). We also use \texttt{universal-ner} for more complex multilingual task which has medium-resource data with non-parallel data.  In this experiment, these data comprises of languages that are divided into \texttt{unseen lang} and \texttt{seen lang} to simulate zero-shot cross-lingual scenarios. Table \ref{tab:data-stat} shows the corresponding datasets' data statistics and language partition. The detailed language partition information can be seen in Appendix~\ref{appendix:langpart}.

\paragraph{Model} We used \texttt{transformers} library \cite{wolf-etal-2020-transformers} and the off-the-shelf implementation of \texttt{xlm-roberta}~\cite{roberta}  and \texttt{mdeberta}~\cite{he2021deberta} models. We used a reduction factor of 2 for the number of student layers compared to the teacher. Additionally, we compared performance by reducing the hidden units by half and keeping the hidden units the same as the teacher's. We experimented with three different model initialization scenarios: copying the weights from \texttt{xlm-roberta-base} or \texttt{mdeberta-v3-base} (\texttt{from-base}), copying from the fine-tuned teacher (\texttt{from-teacher}), and initializing without copying from any model (\texttt{from-scratch}). we compared their performances to understand the differences between these strategies.

\paragraph{Training} To fine-tune the model and perform knowledge distillation, we used AdamW~\cite{loshchilov2018decoupled} as the optimizer, with the default hyperparameters stated in the \texttt{transformers} library. We set the number of epochs to 30 and obtained the best results evaluated on the development set using the F1 score metric. The evaluation steps for \texttt{massive} are as follows: 5000 steps for \S\ref{sec: exp-multi} and \S\ref{sec: exp-cros-lingual}, and 100 steps for \S\ref{sec: exp-en-only}. For \texttt{tsm} and \texttt{universal-ner}, we set the evaluation steps to 500, 250, and 60 for \S\ref{sec: exp-multi}, \S\ref{sec: exp-cros-lingual}, and \S\ref{sec: exp-en-only}, respectively. These differences are due to the data sizes used in the corresponding experiments. The rest of the hyperparameters follow the default configuration in the \texttt{transformers} library. We used an A100 GPU to train our models to run our experiments, running each model's training three times with different seeds. Performances are evaluated using Macro F1-Score. For \texttt{universal-ner}, it is evaluated using \texttt{overall-f1} of \texttt{seqeval}  wrapped in \texttt{evaluate} package\footnote{\url{https://huggingface.co/spaces/evaluate-metric/seqeval}}. 

\section{Multilingual Transferability in KD
} \label{sec: head-multi-kd-exp}

The ability of knowledge distillation (KD) to transfer knowledge across multiple languages efficiently remains unexplored.  As mentioned in \S\ref{sec: methodology}, two components need to be analyzed: model initialization and the distillation process itself. It is still unclear which of these factors contributes the most to the overall performance. Also, in multilingual scenarios, we often encounter situations where not all languages are covered in the training set. Understanding whether KD can facilitate zero-shot cross-lingual (ZSCL) generalization and effectively transfer multilingual knowledge remains unexplored. Building a multilingual dataset is tedious, thus leading researchers to opt for using one language. As a result, it is desirable if such setup can achieve cross-lingual generalization~\cite{artetxe-schwenk-2019-massively}.

\subsection{Weight Copy Transfers More Information vs Distillation Loss} \label{sec: exp-multi}

 \begin{table*}
\centering
\small
\setlength{\tabcolsep}{4pt}
\begin{tabular}{@{}lcccccccc@{}}
\toprule
\multirow{2}{*}{\textbf{Initialization}} & \multicolumn{3}{c}{\textbf{XLM-R}} & \multicolumn{3}{c}{\textbf{mDeBERTa}} & \multirow{2}{*}{\textbf{Avg}} \\
\cmidrule(lr){2-4} \cmidrule(lr){5-7}
& \texttt{massive} & \texttt{tsm} & \texttt{universal-ner} & \texttt{massive} & \texttt{tsm} & \texttt{universal-ner} \\
\midrule
from-base & 80.37 & 60.17 & 78.64 & 80.61 & 58.24 & 77.04 & 72.51 \\
\quad +KD & 81.61 & 65.94 & \textbf{80.29} & 81.82 & 59.11 & \textbf{79.24} & 74.67 \\
\midrule
from-teacher & 81.18 & 62.99 & 79.50 & 80.45 & 58.29 & 78.20 & 73.44 \\
\quad +KD & \textbf{81.63} & \textbf{67.61} & 80.18 & \textbf{82.31} & \textbf{61.08} & 79.17 & \textbf{75.33} \\
\bottomrule
\end{tabular}
\caption{Performance comparison of XLMR and MDEBERTA models on \texttt{from-base} and \texttt{from-teacher} initialization. Bold denotes best methods for each data and their average.}
\label{tab:comparative-performance-base-teacher}
\end{table*}

\begin{table*}
\centering
\small
\setlength{\tabcolsep}{4pt}
\begin{tabular}{@{}llccccccc@{}}
\toprule
\multirow{2}{*}{\textbf{Model}} & \multirow{2}{*}{\textbf{Hidden Size}} & \multicolumn{3}{c}{\textbf{XLM-R}} & \multicolumn{3}{c}{\textbf{mDeBERTa}} & \multirow{2}{*}{\textbf{Avg}} \\
\cmidrule(lr){3-5} \cmidrule(lr){6-8}
& & \texttt{massive} & \texttt{tsm} & \texttt{universal-ner} & \texttt{massive} & \texttt{tsm} & \texttt{universal-ner} & \\
\midrule
\multirow{2}{*}{\texttt{from-scratch}} & 384 & 78.05 & 51.63 & 36.46 & 75.93 & 49.97 & 44.33 & 56.09 \\
& 768 & 79.23 & 54.13 & 48.29 & 76.22 & 49.86 & 52.13 & 59.98 \\
\midrule
\multirow{2}{*}{\texttt{from-teacher}} & 384 & 79.90 & 58.34 & 44.87 & 80.10 & 54.75 & 44.87 & 60.47 \\
& 768 & 81.63 & 67.61 & 80.18 & 82.31 & 61.08 & 79.17 & 75.33 \\
\bottomrule
\end{tabular}
\caption{Comparative performance of XLM-R and mDeBERTa models across different datasets, initialization methods, and hidden sizes using Knowledge Distillation. \texttt{from-scratch} and \texttt{from-teacher} use a layer reduction factor of 2.}
\label{tab:model-performance-hidden-size}
\end{table*}

Table \ref{tab:main-table-result} presents a comparative analysis of model performance with and without the copy-weight initialization strategy  using all languages in the training dataset. The results demonstrate that the \texttt{from-teacher} approach significantly outperforms the \texttt{from-scratch} method, particularly when trained on the \texttt{tsm} and \texttt{universal-ner} datasets. While the performance gap is less pronounced for the \texttt{massive} dataset, it still favors the \texttt{from-teacher} method. These findings suggest that directly copying weights from a larger model can serve as an effective initialization strategy, especially for low-resource scenarios.

Furthermore, the application of knowledge distillation consistently yields performance improvements across all configurations. However, it is noteworthy that the initial weight initialization plays a more critical role in determining overall model performance. The \texttt{from-scratch} initialization, even with knowledge distillation, struggles to match the performance levels achieved by the \texttt{from-teacher}.

To evaluate the effectiveness of different initialization strategies, we compared models initialized from the teacher to those initialized from the base model. As shown in Table \ref{tab:comparative-performance-base-teacher}, the \texttt{from-teacher} approach demonstrates a slight performance advantage over \texttt{from-base}, which can be attributed to the teacher model's prior fine-tuning. The application of Knowledge Distillation (KD) improves performance for both initialization methods. For datasets such as \texttt{massive} and \texttt{universal-ner}, the performance gains are comparable between the two initialization strategies when KD is applied. 

Our previous experiments simply reduced the model size by reducing the layer size while retaining the unit size. In practice, however, we might also want to reduce the unit size. In \texttt{from-teacher} initialization, the performance in all datasets falls significantly in \texttt{tsm} and \texttt{universal-ner} by halving the unit size. This performance reduction is expected due to the model's decreased capacity to retain multilingual information. Additionally, the current approach of uniformly copying weights~\cite{xu2024initializing} may not be optimal for this multilingual distillation task.

\subsection{Knowledge of Unseen Languages is Transferrable with Seen Language Teacher Weight Copy} \label{sec: exp-cros-lingual}

To test the zero-shot cross-lingual performance, we observe two conditions: 1) using the \texttt{seen lang} subset as training data for both the student and teacher models, and 2) using the \texttt{seen lang} subset as training data for the teacher, then using the English language to fine-tune the student model. The motivation is to observe if the model retains multilingual information from the copy-weight, even when fine-tuned using only English.
\begin{table*}
\centering
\small
\setlength{\tabcolsep}{3pt}
\begin{tabular}{@{}l>{\raggedright\arraybackslash}p{2cm}lcccccccc@{}}
\toprule
\multirow{2}{*}{\textbf{Model}} & \multirow{2}{*}{\textbf{Initialization}} & \multirow{2}{*}{\textbf{Train Data}} & \multicolumn{3}{c}{\textbf{XLM-R}} & \multicolumn{3}{c}{\textbf{mDeBERTa}} & \multirow{2}{*}{\textbf{Avg}} \\
\cmidrule(lr){4-6} \cmidrule(lr){7-9}
& & & \texttt{massive} & \texttt{tsm} & \texttt{universal-ner} & \texttt{massive} & \texttt{tsm} & \texttt{universal-ner} & \\
\midrule
Teacher & \texttt{from-base} & \texttt{seen-lang} & 68.22 & 57.11 & 84.38 & 68.54 & 57.04 & 87.50 & 70.47 \\
\midrule
\multirow{4}{*}{Student} & \texttt{from-teacher} & \texttt{seen-lang} & 65.74 & 54.02 & 78.47 & 55.62 & 47.35 & 72.77 & 62.33 \\
& \texttt{from-scratch} & \texttt{seen-lang} & 15.10 & 40.27 & 14.61 & 16.36 & 32.74 & 14.61 & 22.28 \\
\cmidrule{2-10}
& \texttt{from-teacher} & \texttt{english} & 59.65 & 53.35 & 73.73 & 32.82 & 46.18 & 66.61 & 55.39 \\
& \texttt{from-scratch} & \texttt{english} & 7.55 & 33.24 & 7.09 & 3.92 & 28.89 & 7.50 & 14.70 \\
\bottomrule
\end{tabular}
\caption{Cross-lingual performance (F1-scores \%) for XLM-R and mDeBERTa models with different initialization and training data. Knowledge Distillation is used on these models. \textbf{Teacher models are trained using \texttt{seen-lang}}.}
\label{tab:cross-lingual-performance-new}
\end{table*}

Table~\ref{tab:cross-lingual-performance-new} shows the results of zero-shot cross-lingual generalization. The teacher's accuracy drops significantly compared to the scenario in \S\ref{sec: exp-multi}. When using the \texttt{seen lang} subset for training, we observe similar behavior to the previous results, with a slight difference between in with and without KD in both datasets, unlike the results shown in \S\ref{sec: exp-multi}. However, without weight-copy, the performance of each data plummets to near-random answers. This shows that weight-copy preserves multilingual knowledge and enables zero-shot cross-lingual generalization in both high and low-resource scenarios.

Using English as the training data deteriorates the performance in \texttt{massive}'s without KD setup, yet it gains considerable performance by using KD in the copy-weight initialization. Even when using only the English language, the student still retains the ZSCL generalization performance, albeit with reduced effectiveness, which further strengthens our claim regarding copy-weight multilingual generalization.

On the other hand, \texttt{tsm} and \texttt{universal-ner} perform similarly to the student model trained on \texttt{seen lang}, where using KD yields only a marginal increase. This is attributed to KD needing a sufficient amount of data to be effective.

\subsection{Multilingual Distillation is Possible Even if Only English Data is Available} \label{sec: exp-en-only}

We fine-tune the teacher and student models using only English, as this language is the most widely available. Note that, unlike the experiment done in \S\ref{sec: exp-cros-lingual}, we do not train the teacher model with \texttt{seen lang}; we use English instead. We then evaluate it on \texttt{unseen lang} to make the results comparable with those in \S\ref{sec: exp-cros-lingual}. We focus on KD since it has shown a consistent pattern in the previous experiments in \S\ref{sec: exp-multi} and \S\ref{sec: exp-cros-lingual}.

\begin{table}
\centering
\small
\setlength{\tabcolsep}{3pt}
\begin{tabular}{@{}llcccc@{}}
\toprule
\textbf{Model} & \textbf{Initialization} & \rotatebox{90}{\texttt{massive}} & \rotatebox{90}{\texttt{tsm}} & \rotatebox{90}{\texttt{universal-ner}} & \textbf{Avg} \\
\midrule
XLMR (T) & \texttt{from-base} & 56.42 & 58.97 & 77.86 & 64.42 \\
XLMR (S) & \texttt{from-teacher} & 47.85 & 54.18 & 69.05 & 57.03 \\
XLMR (S) & \texttt{from-scratch} & 7.50 & 35.10 & 9.23 & 17.28 \\
\midrule
mDEBERTa (T) & \texttt{from-base} & 63.19 & 59.29 & 77.93 & 66.80 \\
mDEBERTa (S) & \texttt{from-teacher} & 26.83 & 44.54 & 61.97 & 44.45 \\
mDEBERTa (S) & \texttt{from-scratch} & 3.72 & 28.45 & 10.12 & 14.10 \\
\bottomrule
\end{tabular}
\caption{Cross-lingual performance (F1-scores). (T) \textbf{denotes Teacher model trained on English only}; (S) denotes Student model.}
\label{tab:exp-en-only}
\end{table}

Table \ref{tab:exp-en-only} shows the results of the current experiment. Compared to using a fine-tuned teacher model with \texttt{seen lang}, we can see that \texttt{massive} performance dropped by about 12\%, while it slightly improved for \texttt{tsm} and \texttt{universal-ner}. We argue that \texttt{tsm} and \texttt{universal-ner} data is non-parallel and contains substantial fewer instances than \texttt{massive} . As a result, these performances do not follow the same pattern as \texttt{massive}. 

Although \texttt{from-teacher} exhibited the highest score, there is a significant gap compared to the \S\ref{sec: exp-cros-lingual} experiment. Having one language trained on the teacher makes copy-weight initialization less effective, yet the model still retains some multilingual capability. In contrast, \texttt{from-scratch} performs similarly and near random score\footnote{Random score is obtained by making all predictions equal to the major class in the dataset.}.

\section{Behavior Analysis in Copy-weight Strategy}

In \S\ref{sec: head-multi-kd-exp}, we summarized that model initialization strategy significantly impacts transferring multilingual knowledge, with \texttt{from-teacher} performing the best. This section provides more detailed analysis related to the model's characteristics when using the copy-weight strategy, such as zero-shot copy classification performance (\S\ref{sec: exp-zero-shot}), training speed after the weight is copied from the teacher to the student model across different data subsets (\S\ref{sec: exp-less-data}), and 3) performance across different data subsets (\S\ref{sec: converge-faster}).

The experiments performed in this section will use KD and the setup described in \S\ref{sec: exp-multi}, with full hidden size. We focus on analyzing the behavior of the copy-weight strategy.

\subsection{Weight Copy model preserve some information even without finetuning} \label{sec: exp-zero-shot}

\begin{table}
\centering
\small
\begin{tabular}{llll}
\toprule
Training Method & \texttt{massive} & \texttt{tsm} & \texttt{universal-ner} \\
\midrule
With Finetune & 81.63 & 67.61 &  80.18 \\
Without Finetune & 38.05 & 33.57 & 7.79  \\ 
Random Score & 7.02 & 33.33 & 1.75 \\
\bottomrule

\end{tabular}
\caption{Zero-shot performance by only copying the weight of the respective fine-tuned teacher to their half-layer students. Random score obtained on \texttt{universal-ner} is generated through average of random predictions in 30 runs.}
\label{tab: zero-shot-score}

\end{table}

Given the that copy-weight approach is better than the distillation technique itself, we investigate how much multilingual information is retained simply by copying the weights without any additional fine-tuning. Table \ref{tab: zero-shot-score} provides the performance results. We observe that these scores are substantially lower than when fine-tuning is performed. Intriguingly, \texttt{massive} and \texttt{universal-ner} scores are not as low as random guesses, implying that some knowledge is still retained, though not fully 'connected,' and needs to be fine-tuned. On the other hand, \texttt{tsm} shows performance comparable to random guessing. We hypothesize that this is due to the low number of instances in \texttt{tsm}, which do not preserve the inherent bias of multilingual knowledge as strongly as the others.

\subsection{Weight Copy Models Achieve Higher Performance with Less Data} \label{sec: exp-less-data}

The experiment in \S\ref{sec: exp-multi} demonstrated that the copy-weight approach exhibited higher performance, especially in the \texttt{massive} dataset due to its large number of instances. We argue that these results are attributed to better initialization, which enhances data efficiency. To test this hypothesis, we conducted an experiment by creating four subsets of the \texttt{massive} dataset, consisting of 1\%, 5\%, 10\%, and 20\% of the original data. These subsets were generated using stratified sampling based on the label distribution for each language.

\begin{figure}
    \centering
    \includegraphics[width=1\linewidth]{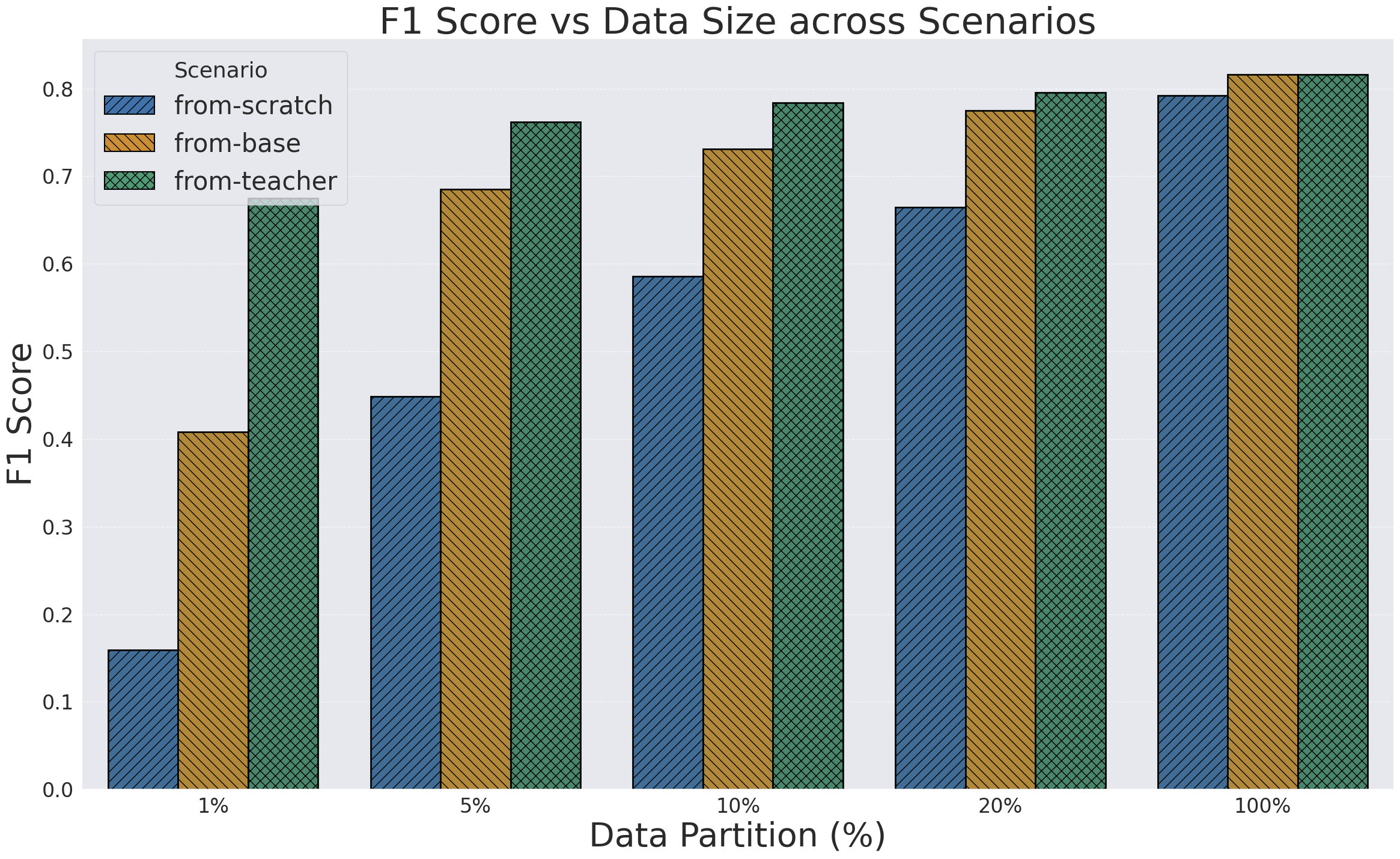}
    \caption{Performance across different data subsets in different initialization strategies.}
   \label{fig: score-partition}
\end{figure}

Figure \ref{fig: score-partition} illustrates the results for the three model initialization strategies. We observe a pattern where using more data corresponds to higher scores. The performance order is consistent, with the best scores achieved by \texttt{from-teacher} and the worst by \texttt{from-scratch}. In the 1\% data subset, \texttt{from-teacher} achieved around 69\% f1-score, showing a significant gap compared to the others, with more than a 20\% difference. However, as the dataset size increases, the gap between scenarios becomes smaller, yet \texttt{from-teacher} consistently exhibits the best results. This demonstrates that utilizing the teacher's fine-tuned weights, even in a low-resource setting, benefits from the inherited information, providing better scores.

\subsection{Weight Copy Provides Better Initialization -- Model Converged Faster} \label{sec: converge-faster}


\begin{figure}
    \centering
    \includegraphics[width=1\linewidth]{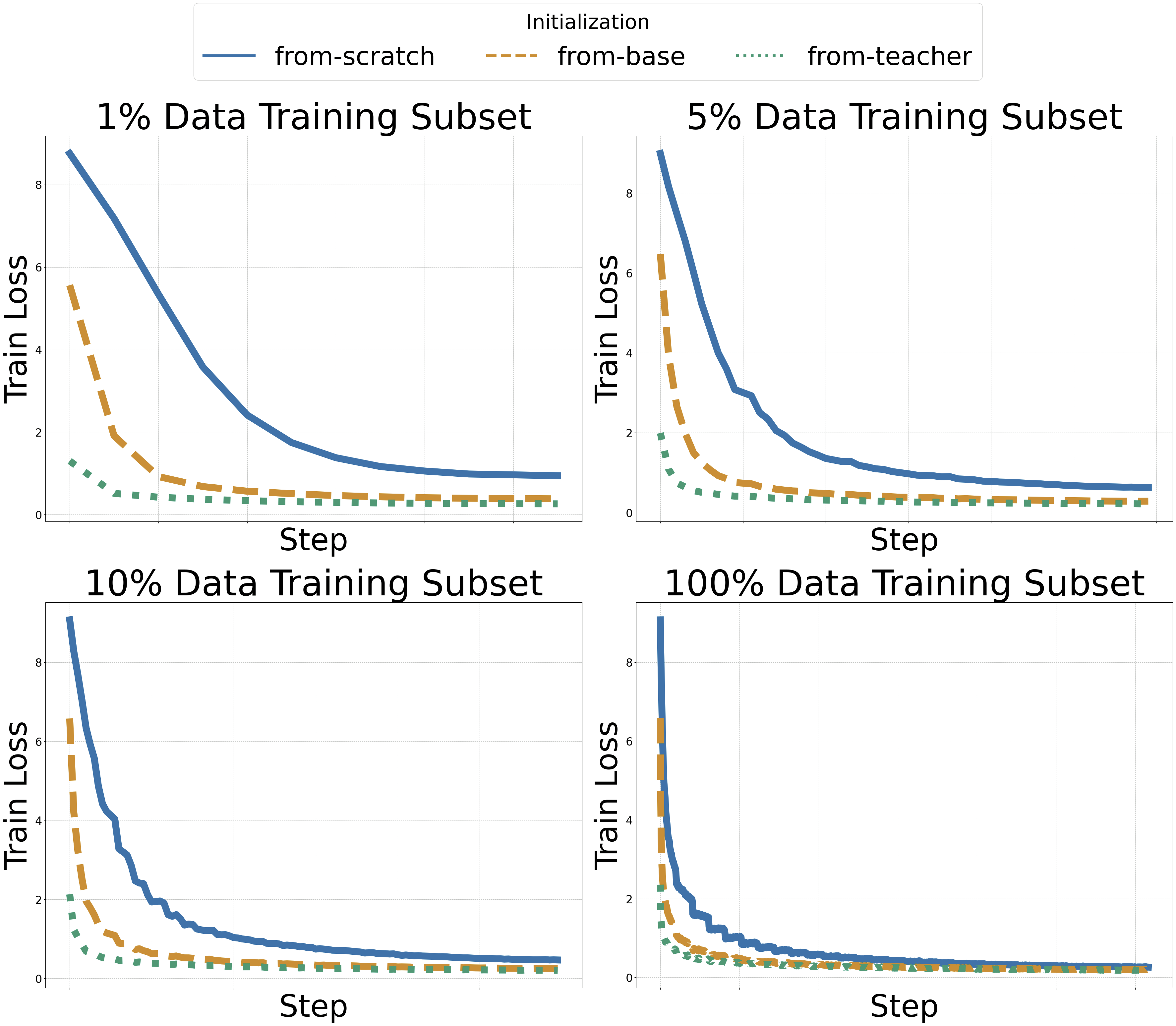}
    \caption{Training loss plot per step across different data subsets.}
    \label{fig: loss-data-partition}
\end{figure}

So far, we have explored that different data subsets exhibit different performances across model initialization strategies. This might also correlate with the training speed due to a better start. Thus, we are also interested in exploring learning efficiency by comparing the learning speed of each strategy with each data subset.

The resulting score correlates with the learning speed, depicted in Figure \ref{fig: loss-data-partition}. Using the full data subset, we observe that the order of scenarios sorted by learning speed is similar to the order in Figure \ref{fig: score-partition}. With smaller data subsets, the gap in training loss between each model is wide, with \texttt{from-teacher} showing the fastest convergence rate. As more data is added, the gap becomes smaller. We posit that this is attributed from the fine-tuned teacher weight-copy, making the model learns faster. Furthermore,  this also indicates that copying the teacher's weights in low-resource settings not only improves the score but also accelerates learning speed, reducing the cost of training the model.

\section{Related Work} \label{sec: related-work}

\paragraph{Model Initialization} Model initialization is crucial when training a model. \citet{glorot2010understanding} introduced a method to properly initialize the weights of neural networks using a normal distribution to avoid issues related to vanishing and exploding gradients during training. This approach has been extended by several others, such as \citet{he2015delving}, \citet{mishkin2016all}, and \citet{saxe2014exact}, to add robustness to gradient problems. While these methods address numerical instability, they do not incorporate inherent initial knowledge. Transfer learning~\cite{zhuang2020comprehensive, howard-ruder-2018-universal} provides a way to start training a model with better initialization with prior knowledge. We pre-train the model on unlabeled data and then fine-tune it on the desired task. Multilingual models like DeBERTa~\cite{he2021deberta}, mBERT~\cite{devlin-etal-2019-bert}, and XLM-R~\cite{roberta} can be used to train models that handle multiple languages. However, these models require extensive training resources to create. 

\paragraph{Knowledge Distillation} Knowledge distillation (KD) \citep{hinton2015distil} produces models with fewer parameters (student model) guided by a larger model (teacher model), often resulting in higher quality than models trained from scratch. In NLP, KD can be applied directly to task-specific or downstream tasks \citep{nityasya2022student, adhikari-etal-2020-exploring, Liu_2020}, or during the pre-training phase of the student model \cite{sun-etal-2020-mobilebert}, which can then be fine-tuned. Several works apply KD during both pre-training and fine-tuning steps \citep{jiao-etal-2020-tinybert, sanh2019distilbert, liu-etal-2020-fastbert}. The aspects of the teacher model that the student should mimic can vary; a common approach is for the student to mimic only the probability distribution of the teacher's prediction layer output. However, \citet{jiao-etal-2020-tinybert} and \citet{sun-etal-2020-mobilebert} also include outputs such as the teacher model's layer outputs, attention layers, and embedding layers. \citet{wang-etal-2020-structure} and \citet{ansell-etal-2023-distilling} explore the potential of KD in multilingual settings, with the latter utilizing sparse fine-tuning and a setup similar to \citet{jiao-etal-2020-tinybert}. However, these works do not thoroughly investigate the behavior of the approach, such as the impact of initialization and data size. This research work dives into probing the influence of these components, providing better insight into multilingual performance capability using weight-copy techniques.

\section{Conclusion} \label{sec: conclusion}

In this work, we observed the effectiveness of Knowledge Distillation, in multilingual settings, focusing on identifying the factors that significantly influence the performance of student models. Our findings demonstrated that model initialization, specifically through weight copying from a fine-tuned teacher model, plays a crucial role in enhancing the performance and learning speed of student models. This finding was consistent across both high-resource and low-resource datasets, highlighting the importance of weight initialization in retaining multilingual knowledge and facilitating effective KD. 

These insights underscore the critical role of initialization in KD, suggesting that simple yet effective strategies, such as weight copying, can lead to substantial performance gains without requiring extensive data or computational resources. This work contributes valuable and practical insights to developing efficient and high-performing multilingual models, particularly in resource-constrained environments. A promising future work is to propose a novel, efficient initialization method that avoids the need for any expensive step, such as pre-training step, in preparing the student model.

\section{Limitations}


In this work, we focus on sequence classification and token classification tasks, which may not generalize to other tasks, such as Natural Language Generation.  The languages observed in this work are those represented in \texttt{massive}, \texttt{universal-ner} and \texttt{tsm}, which do not include every possible language. Additionally, our study is limited to specific model sizes and architectures (e.g., XLM-RoBERTa and mDEBERTa). 

We concentrate on analyzing multilingual capability, specifically zero-shot fine-tuning generalization, rather than zero-shot inference as exhibited in Large Language Models. This focus is due to the amount of computational cost associated with fine-tuning such large models.

Finally, while using unlabeled datasets for distillation may improve the system's performance, it adds another layer of complexity to our work. Analyzing the data for use in a multilingual setting is beyond the scope of this study. We leave this for future work.


\section*{Ethical Considerations}

This work has no ethical issues, as it focuses on analyzing the inner workings of a multilingual model in knowledge distillation. All artifacts used in this research are permitted for research purposes and align with their intended usage in multilingualism. Additionally, the data utilized does not contain any personally identifiable information or offensive content.

\bibliography{anthology,custom}

\begin{thebibliography}{31}
\expandafter\ifx\csname natexlab\endcsname\relax\def\natexlab#1{#1}\fi

\bibitem[{Adhikari et~al.(2020)Adhikari, Ram, Tang, Hamilton, and Lin}]{adhikari-etal-2020-exploring}
Ashutosh Adhikari, Achyudh Ram, Raphael Tang, William~L. Hamilton, and Jimmy Lin. 2020.
\newblock \href {https://doi.org/10.18653/v1/2020.repl4nlp-1.10} {Exploring the limits of simple learners in knowledge distillation for document classification with {D}oc{BERT}}.
\newblock In \emph{Proceedings of the 5th Workshop on Representation Learning for NLP}, pages 72--77, Online. Association for Computational Linguistics.

\bibitem[{Aji et~al.(2022)Aji, Winata, Koto, Cahyawijaya, Romadhony, Mahendra, Kurniawan, Moeljadi, Prasojo, Baldwin, Lau, and Ruder}]{aji-etal-2022-one}
Alham~Fikri Aji, Genta~Indra Winata, Fajri Koto, Samuel Cahyawijaya, Ade Romadhony, Rahmad Mahendra, Kemal Kurniawan, David Moeljadi, Radityo~Eko Prasojo, Timothy Baldwin, Jey~Han Lau, and Sebastian Ruder. 2022.
\newblock \href {https://doi.org/10.18653/v1/2022.acl-long.500} {One country, 700+ languages: {NLP} challenges for underrepresented languages and dialects in {I}ndonesia}.
\newblock In \emph{Proceedings of the 60th Annual Meeting of the Association for Computational Linguistics (Volume 1: Long Papers)}, pages 7226--7249, Dublin, Ireland. Association for Computational Linguistics.

\bibitem[{Ansell et~al.(2023)Ansell, Ponti, Korhonen, and Vuli{\'c}}]{ansell-etal-2023-distilling}
Alan Ansell, Edoardo~Maria Ponti, Anna Korhonen, and Ivan Vuli{\'c}. 2023.
\newblock \href {https://doi.org/10.18653/v1/2023.findings-acl.517} {Distilling efficient language-specific models for cross-lingual transfer}.
\newblock In \emph{Findings of the Association for Computational Linguistics: ACL 2023}, pages 8147--8165, Toronto, Canada. Association for Computational Linguistics.

\bibitem[{Artetxe and Schwenk(2019)}]{artetxe-schwenk-2019-massively}
Mikel Artetxe and Holger Schwenk. 2019.
\newblock \href {https://doi.org/10.1162/tacl_a_00288} {Massively multilingual sentence embeddings for zero-shot cross-lingual transfer and beyond}.
\newblock \emph{Transactions of the Association for Computational Linguistics}, 7:597--610.

\bibitem[{Barbieri et~al.(2022)Barbieri, Espinosa~Anke, and Camacho-Collados}]{barbieri-etal-2022-xlm}
Francesco Barbieri, Luis Espinosa~Anke, and Jose Camacho-Collados. 2022.
\newblock \href {https://aclanthology.org/2022.lrec-1.27} {{XLM}-{T}: Multilingual language models in {T}witter for sentiment analysis and beyond}.
\newblock In \emph{Proceedings of the Thirteenth Language Resources and Evaluation Conference}, pages 258--266, Marseille, France. European Language Resources Association.

\bibitem[{Conneau et~al.(2020)Conneau, Khandelwal, Goyal, Chaudhary, Wenzek, Guzm{\'a}n, Grave, Ott, Zettlemoyer, and Stoyanov}]{conneau-etal-2020-unsupervised}
Alexis Conneau, Kartikay Khandelwal, Naman Goyal, Vishrav Chaudhary, Guillaume Wenzek, Francisco Guzm{\'a}n, Edouard Grave, Myle Ott, Luke Zettlemoyer, and Veselin Stoyanov. 2020.
\newblock \href {https://doi.org/10.18653/v1/2020.acl-main.747} {Unsupervised cross-lingual representation learning at scale}.
\newblock In \emph{Proceedings of the 58th Annual Meeting of the Association for Computational Linguistics}, pages 8440--8451, Online. Association for Computational Linguistics.

\bibitem[{Devlin et~al.(2019)Devlin, Chang, Lee, and Toutanova}]{devlin-etal-2019-bert}
Jacob Devlin, Ming-Wei Chang, Kenton Lee, and Kristina Toutanova. 2019.
\newblock \href {https://doi.org/10.18653/v1/N19-1423} {{BERT}: Pre-training of deep bidirectional transformers for language understanding}.
\newblock In \emph{Proceedings of the 2019 Conference of the North {A}merican Chapter of the Association for Computational Linguistics: Human Language Technologies, Volume 1 (Long and Short Papers)}, pages 4171--4186, Minneapolis, Minnesota. Association for Computational Linguistics.

\bibitem[{FitzGerald et~al.(2022)FitzGerald, Hench, Peris, Mackie, Rottmann, Sanchez, Nash, Urbach, Kakarala, Singh, Ranganath, Crist, Britan, Leeuwis, Tur, and Natarajan}]{fitzgerald2022massive}
Jack FitzGerald, Christopher Hench, Charith Peris, Scott Mackie, Kay Rottmann, Ana Sanchez, Aaron Nash, Liam Urbach, Vishesh Kakarala, Richa Singh, Swetha Ranganath, Laurie Crist, Misha Britan, Wouter Leeuwis, Gokhan Tur, and Prem Natarajan. 2022.
\newblock \href {http://arxiv.org/abs/2204.08582} {Massive: A 1m-example multilingual natural language understanding dataset with 51 typologically-diverse languages}.

\bibitem[{Glorot and Bengio(2010)}]{glorot2010understanding}
Xavier Glorot and Yoshua Bengio. 2010.
\newblock Understanding the difficulty of training deep feedforward neural networks.
\newblock In \emph{Proceedings of the Thirteenth International Conference on Artificial Intelligence and Statistics (AISTATS)}, volume~9, pages 249--256. JMLR Workshop and Conference Proceedings.

\bibitem[{He et~al.(2015)He, Zhang, Ren, and Sun}]{he2015delving}
Kaiming He, Xiangyu Zhang, Shaoqing Ren, and Jian Sun. 2015.
\newblock Delving deep into rectifiers: Surpassing human-level performance on imagenet classification.
\newblock \emph{Proceedings of the IEEE international conference on computer vision (ICCV)}, pages 1026--1034.

\bibitem[{He et~al.(2021)He, Liu, Gao, and Chen}]{he2021deberta}
Pengcheng He, Xiaodong Liu, Jianfeng Gao, and Weizhu Chen. 2021.
\newblock \href {http://arxiv.org/abs/2006.03654} {Deberta: Decoding-enhanced bert with disentangled attention}.

\bibitem[{Hinton et~al.(2015)Hinton, Vinyals, and Dean}]{hinton2015distil}
Geoffrey Hinton, Oriol Vinyals, and Jeffrey Dean. 2015.
\newblock \href {http://arxiv.org/abs/1503.02531} {Distilling the knowledge in a neural network}.
\newblock In \emph{NIPS Deep Learning and Representation Learning Workshop}.

\bibitem[{Howard and Ruder(2018)}]{howard-ruder-2018-universal}
Jeremy Howard and Sebastian Ruder. 2018.
\newblock \href {https://doi.org/10.18653/v1/P18-1031} {Universal language model fine-tuning for text classification}.
\newblock In \emph{Proceedings of the 56th Annual Meeting of the Association for Computational Linguistics (Volume 1: Long Papers)}, pages 328--339, Melbourne, Australia. Association for Computational Linguistics.

\bibitem[{Jiao et~al.(2020)Jiao, Yin, Shang, Jiang, Chen, Li, Wang, and Liu}]{jiao-etal-2020-tinybert}
Xiaoqi Jiao, Yichun Yin, Lifeng Shang, Xin Jiang, Xiao Chen, Linlin Li, Fang Wang, and Qun Liu. 2020.
\newblock \href {https://doi.org/10.18653/v1/2020.findings-emnlp.372} {{T}iny{BERT}: Distilling {BERT} for natural language understanding}.
\newblock In \emph{Findings of the Association for Computational Linguistics: EMNLP 2020}, pages 4163--4174, Online. Association for Computational Linguistics.

\bibitem[{Liu et~al.(2020{\natexlab{a}})Liu, Zhou, Wang, Zhao, Deng, and Ju}]{liu-etal-2020-fastbert}
Weijie Liu, Peng Zhou, Zhiruo Wang, Zhe Zhao, Haotang Deng, and Qi~Ju. 2020{\natexlab{a}}.
\newblock \href {https://doi.org/10.18653/v1/2020.acl-main.537} {{F}ast{BERT}: a self-distilling {BERT} with adaptive inference time}.
\newblock In \emph{Proceedings of the 58th Annual Meeting of the Association for Computational Linguistics}, pages 6035--6044, Online. Association for Computational Linguistics.

\bibitem[{Liu et~al.(2019)Liu, Ott, Goyal, Du, Joshi, Chen, Levy, Lewis, Zettlemoyer, and Stoyanov}]{roberta}
Yinhan Liu, Myle Ott, Naman Goyal, Jingfei Du, Mandar Joshi, Danqi Chen, Omer Levy, Mike Lewis, Luke Zettlemoyer, and Veselin Stoyanov. 2019.
\newblock Roberta: A robustly optimized bert pretraining approach.
\newblock \emph{arXiv preprint arXiv:1907.11692}.

\bibitem[{Liu et~al.(2020{\natexlab{b}})Liu, Zhang, and Wang}]{Liu_2020}
Yuang Liu, Wei Zhang, and Jun Wang. 2020{\natexlab{b}}.
\newblock \href {https://doi.org/10.1016/j.neucom.2020.07.048} {Adaptive multi-teacher multi-level knowledge distillation}.
\newblock \emph{Neurocomputing}, 415:106--113.

\bibitem[{Loshchilov and Hutter(2019)}]{loshchilov2018decoupled}
Ilya Loshchilov and Frank Hutter. 2019.
\newblock \href {https://openreview.net/forum?id=Bkg6RiCqY7} {Decoupled weight decay regularization}.
\newblock In \emph{International Conference on Learning Representations}.

\bibitem[{Mayhew et~al.(2024)Mayhew, Blevins, Liu, Šuppa, Gonen, Imperial, Karlsson, Lin, Ljubešić, Miranda, Plank, Riab, and Pinter}]{mayhew2024universal}
Stephen Mayhew, Terra Blevins, Shuheng Liu, Marek Šuppa, Hila Gonen, Joseph~Marvin Imperial, Börje~F. Karlsson, Peiqin Lin, Nikola Ljubešić, LJ~Miranda, Barbara Plank, Arij Riab, and Yuval Pinter. 2024.
\newblock \href {https://aclanthology.org/2024.naacl-long.243/} {Universal ner: A gold-standard multilingual named entity recognition benchmark}.
\newblock In \emph{Proceedings of the 2024 Conference of the North American Chapter of the Association for Computational Linguistics (NAACL)}.

\bibitem[{Mishkin and Matas(2016)}]{mishkin2016all}
Dmytro Mishkin and Jiri Matas. 2016.
\newblock All you need is a good init.
\newblock In \emph{International Conference on Learning Representations (ICLR)}.

\bibitem[{Nityasya et~al.(2022)Nityasya, Wibowo, Chevi, Prasojo, and Aji}]{nityasya2022student}
Made~Nindyatama Nityasya, Haryo~Akbarianto Wibowo, Rendi Chevi, Radityo~Eko Prasojo, and Alham~Fikri Aji. 2022.
\newblock \href {http://arxiv.org/abs/2201.00558} {Which student is best? a comprehensive knowledge distillation exam for task-specific bert models}.

\bibitem[{Nityasya et~al.(2021)Nityasya, Wibowo, Prasojo, and Aji}]{nityasya2021costs}
Made~Nindyatama Nityasya, Haryo~Akbarianto Wibowo, Radityo~Eko Prasojo, and Alham~Fikri Aji. 2021.
\newblock \href {http://arxiv.org/abs/2012.08958} {Costs to consider in adopting nlp for your business}.

\bibitem[{Ruder(2022)}]{ruder2022statemultilingualai}
Sebastian Ruder. 2022.
\newblock {The State of Multilingual AI}.
\newblock \url{http://ruder.io/state-of-multilingual-ai/}.

\bibitem[{Sanh et~al.(2019)Sanh, Debut, Chaumond, and Wolf}]{sanh2019distilbert}
Victor Sanh, Lysandre Debut, Julien Chaumond, and Thomas Wolf. 2019.
\newblock Distilbert, a distilled version of bert: smaller, faster, cheaper and lighter.
\newblock In \emph{$NeurIPS EMC^2 Workshop$}.

\bibitem[{Sanh et~al.(2020)Sanh, Debut, Chaumond, and Wolf}]{sanh2020distilbert}
Victor Sanh, Lysandre Debut, Julien Chaumond, and Thomas Wolf. 2020.
\newblock \href {http://arxiv.org/abs/1910.01108} {Distilbert, a distilled version of bert: smaller, faster, cheaper and lighter}.

\bibitem[{Saxe et~al.(2014)Saxe, McClelland, and Ganguli}]{saxe2014exact}
Andrew~M Saxe, James~L McClelland, and Surya Ganguli. 2014.
\newblock Exact solutions to the nonlinear dynamics of learning in deep linear neural networks.
\newblock In \emph{International Conference on Learning Representations (ICLR)}.

\bibitem[{Sun et~al.(2020)Sun, Yu, Song, Liu, Yang, and Zhou}]{sun-etal-2020-mobilebert}
Zhiqing Sun, Hongkun Yu, Xiaodan Song, Renjie Liu, Yiming Yang, and Denny Zhou. 2020.
\newblock \href {https://doi.org/10.18653/v1/2020.acl-main.195} {{M}obile{BERT}: a compact task-agnostic {BERT} for resource-limited devices}.
\newblock In \emph{Proceedings of the 58th Annual Meeting of the Association for Computational Linguistics}, pages 2158--2170, Online. Association for Computational Linguistics.

\bibitem[{Wang et~al.(2020)Wang, Jiang, Bach, Wang, Huang, and Tu}]{wang-etal-2020-structure}
Xinyu Wang, Yong Jiang, Nguyen Bach, Tao Wang, Fei Huang, and Kewei Tu. 2020.
\newblock \href {https://doi.org/10.18653/v1/2020.acl-main.304} {Structure-level knowledge distillation for multilingual sequence labeling}.
\newblock In \emph{Proceedings of the 58th Annual Meeting of the Association for Computational Linguistics}, pages 3317--3330, Online. Association for Computational Linguistics.

\bibitem[{Wolf et~al.(2020)Wolf, Debut, Sanh, Chaumond, Delangue, Moi, Cistac, Rault, Louf, Funtowicz, Davison, Shleifer, von Platen, Ma, Jernite, Plu, Xu, Le~Scao, Gugger, Drame, Lhoest, and Rush}]{wolf-etal-2020-transformers}
Thomas Wolf, Lysandre Debut, Victor Sanh, Julien Chaumond, Clement Delangue, Anthony Moi, Pierric Cistac, Tim Rault, Remi Louf, Morgan Funtowicz, Joe Davison, Sam Shleifer, Patrick von Platen, Clara Ma, Yacine Jernite, Julien Plu, Canwen Xu, Teven Le~Scao, Sylvain Gugger, Mariama Drame, Quentin Lhoest, and Alexander Rush. 2020.
\newblock \href {https://doi.org/10.18653/v1/2020.emnlp-demos.6} {Transformers: State-of-the-art natural language processing}.
\newblock In \emph{Proceedings of the 2020 Conference on Empirical Methods in Natural Language Processing: System Demonstrations}, pages 38--45, Online. Association for Computational Linguistics.

\bibitem[{Xu et~al.(2024)Xu, Chen, Vishniakov, Yin, Shen, Darrell, Liu, and Liu}]{xu2024initializing}
Zhiqiu Xu, Yanjie Chen, Kirill Vishniakov, Yida Yin, Zhiqiang Shen, Trevor Darrell, Lingjie Liu, and Zhuang Liu. 2024.
\newblock \href {https://openreview.net/forum?id=dyrGMhicMw} {Initializing models with larger ones}.
\newblock In \emph{The Twelfth International Conference on Learning Representations}.

\bibitem[{Zhuang et~al.(2020)Zhuang, Qi, Duan, Xi, Zhu, Zhu, Xiong, and He}]{zhuang2020comprehensive}
Fuzhen Zhuang, Zhiyuan Qi, Keyu Duan, Dongbo Xi, Yongchun Zhu, Hengshu Zhu, Hui Xiong, and Qing He. 2020.
\newblock \href {http://arxiv.org/abs/1911.02685} {A comprehensive survey on transfer learning}.

\end{thebibliography}
\bibliographystyle{acl_natbib}

\appendix
\newpage
\onecolumn
\label{sec:appendix}

\begin{table*}
\centering
\small
\setlength{\tabcolsep}{4pt}
\begin{tabular}{@{}llcccccccccc@{}}
\toprule
\multirow{2}{*}{\textbf{Model}} & \multirow{2}{*}{\textbf{Initialization}} & \multirow{2}{*}{\textbf{\#L}} & \multirow{2}{*}{\textbf{\#U}} & \multicolumn{2}{c}{\texttt{massive}} & \multicolumn{2}{c}{\texttt{tsm}} & \multicolumn{2}{c}{\texttt{universal-ner}} \\
\cmidrule(lr){5-6} \cmidrule(lr){7-8} \cmidrule(lr){9-10}
& & & & \textbf{NON} & \textbf{KD} & \textbf{NON} & \textbf{KD} & \textbf{NON} & \textbf{KD} \\
\midrule
\multirow{7}{*}{XLM-R} & 
from-base (T) & 12 & 768 & 80.13 & -- & 70.10 & -- & 87.66 & -- \\
\cmidrule{2-10}
& from-teacher & 6 & 768 & 81.18 & 81.63 & 62.99 & 67.61 & 79.50 & 80.18 \\
& from-base    & 6 & 768 & 80.37 & 81.61 & 60.17 & 65.94 & 78.64 & 80.29 \\
& from-scratch & 6 & 768 & 75.19 & 79.23 & 50.20 & 54.13 & 45.68 & 48.29 \\
\cmidrule{2-10}
& from-teacher & 6 & 384 & 78.20 & 79.90 & 56.74 & 58.34\textbf{ }& 47.15 & 44.87 \\
& from-base    & 6 & 384 & 77.55 & 79.41 & 55.03 & 58.15 & 45.94 & 44.24 \\
& from-scratch & 6 & 384 & 75.27 & 78.05 & 50.01 & 51.63 & 40.80 & 36.46 \\
\midrule
\multirow{7}{*}{mDeBERTa} & 
from-base (T) & 12 & 768 & 81.36 & -- & 66.96 & -- & 88.81 & -- \\
\cmidrule{2-10}
& from-teacher & 6 & 768 & 80.45 & 82.31 & 58.29 & 61.08 & 78.20 & 79.17 \\
& from-base    & 6 & 768 & 80.61 & 81.82 & 58.24 & 59.11 & 77.04 & 79.24 \\
& from-scratch & 6 & 768 & 75.93 & 76.22 & 49.86 & 51.80 & 46.21 & 44.56 \\
\cmidrule{2-10}
& from-teacher & 6 & 384 & 79.76 & 80.10  & 54.59 & 54,75 & 62.57 & 59.72 \\
& from-base    & 6 & 384 & 78.49 & 78.35 & 53.94 & 46.20 & 62.81 & 59.24 \\
& from-scratch & 6 & 384 & 76.22 & 75.93 & 49.97 & 43.39 & 45.55 & 44.33 \\
\bottomrule
\end{tabular}
\caption{F1-scores (\%) for XLM-R and mDeBERTa models with different initializations, both with (KD) and without (NON) knowledge distillation, across three datasets. \#L: Number of Layers, \#U: Number of Units, T: Teacher.}
\label{tab:main-exp}
\end{table*}

\begin{table*}
\centering
\small
\setlength{\tabcolsep}{4pt}
\begin{tabular}{@{}llllcccccc@{}}
\toprule
\textbf{Model} & \textbf{Initialization} & \textbf{Training Data} & \textbf{Test Data} & \multicolumn{2}{c}{\texttt{massive}} & \multicolumn{2}{c}{\texttt{tsm}} & \multicolumn{2}{c}{\texttt{universal-ner}} \\
\cmidrule(lr){5-6} \cmidrule(lr){7-8} \cmidrule(lr){9-10}
& & & & \textbf{NON-KD} & \textbf{KD} & \textbf{NON-KD} & \textbf{KD} & \textbf{NON-KD} & \textbf{KD} \\
\midrule
\multirow{8}{*}{XLM-R} & \texttt{from-base} (T) & \texttt{seen lang} & \texttt{unseen lang} & 68.22 & -- & 57.11 & -- & 84.38
& -- \\
\cmidrule{2-10}
& \texttt{from-teacher} & \texttt{seen lang} & \texttt{unseen lang} & 64.30 & 65.74 & 53.99 & 54.02 & 77.05 & 78.47 \\
& \texttt{from-base}    & \texttt{seen lang} & \texttt{unseen lang} & 62.77 & 64.82 & 52.31 & 52.36 & 77.27 & 79.15 \\
& \texttt{from-scratch} & \texttt{seen lang} & \texttt{unseen lang} & 15.08 & 15.10 & 37.93 & 40.27 & 15.58 & 14.61 \\
\cmidrule{2-10}
& \texttt{from-teacher} & \texttt{english} & \texttt{unseen lang} & 47.23 & 59.65 & 54.36 & 53.35 & 66.96 & 73.73 \\
& \texttt{from-base}    & \texttt{english} & \texttt{unseen lang} & 38.69 & 46.16 & 50.47 & 49.74 & 64.99 & 71.73 \\
& \texttt{from-scratch} & \texttt{english} & \texttt{unseen lang} & 5.25 & 7.55 & 34.39 & 33.24 & 9.21 & 7.09 \\
\midrule
\multirow{8}{*}{mDeBERTa} & \texttt{from-base} (T) & \texttt{seen lang} & \texttt{unseen lang} & 68.54 & -- & 57.04 & -- & 87.50 & -- \\
\cmidrule{2-10}
& \texttt{from-teacher} & \texttt{seen lang} & \texttt{unseen lang} & 50.91& 55.62 &  48.10 & 47.35 & 73.28 & 72.77 \\
& \texttt{from-base}    & \texttt{seen lang} & \texttt{unseen lang} & 47.59& 54.90 & 46.66 & 49.04 & 69.13 & 73.56 \\
& \texttt{from-scratch} & \texttt{seen lang} & \texttt{unseen lang} & 14.80& 16.36& 39.59 & 32.74& 15.58 & 14.61 \\
\cmidrule{2-10}
& \texttt{from-teacher} & \texttt{english} & \texttt{unseen lang} & 20.98& 32.82 & 43.14 & 46.18& 59.10& 66.61\\
& \texttt{from-base}    & \texttt{english} & \texttt{unseen lang} & 16.23& 33.64 & 43.38 & 48.32& 53.37& 65.34\\
& \texttt{from-scratch} & \texttt{english} & \texttt{unseen lang} & 7.09& 3.92& 38.45 & 28.89& 8.26& 7.50\\
\bottomrule
\end{tabular}
\caption{F1-scores (\%) for zero-shot cross-lingual generalization in Knowledge Distillation for XLM-R and mDeBERTa models. We fine-tune the teacher model using \texttt{seen lang} and fine-tune the student model according to the training data provided. NON-KD follows the student model configuration. All models contain 6 layers, except for the teacher models which have 12 layers.}
\label{tab:cross-lingual-result}
\end{table*}

\begin{table}[htbp]
\centering
\small
\begin{tabular}{@{}llllll@{}}
\toprule
\textbf{Code} & \textbf{Language} & \textbf{Code} & \textbf{Language} & \textbf{Code} & \textbf{Language} \\
\midrule
ceb & Cebuano & en & English & sk & Slovak \\
da & Danish & hr & Croatian & sr & Serbian \\
de & German & pt & Portuguese & sv & Swedish \\
ru & Russian & tl & Tagalog & zh & Chinese \\
\bottomrule
\end{tabular}
\caption{Languages in the \texttt{universal-ner} dataset}
\label{tab:universal-ner-languages}
\end{table}

\begin{table}[htbp]
\centering
\small
\begin{tabular}{@{}llllll@{}}
\toprule
\textbf{Code} & \textbf{Language} & \textbf{Code} & \textbf{Language} & \textbf{Code} & \textbf{Language} \\
\midrule
arabic & Arabic & german & German & portuguese & Portuguese \\
english & English & hindi & Hindi & spanish & Spanish \\
french & French & italian & Italian & & \\
\bottomrule
\end{tabular}
\caption{Languages in the \texttt{tsm} dataset}
\label{tab:tsm-languages}
\end{table}

\section{Additional Experiment Results}

A complete and comprehensive experiment results can be seen in Table~\ref{tab:main-exp} and Table~\ref{tab:cross-lingual-result} for all languages training and zero-shot cross-lingual performance in weight-copy, respectively.

\section{Language Partition Explanation} \label{appendix:langpart}

Detail of language partitions in the training dataset can be seen in Table~\ref{tab:language-codes} (\texttt{massive}), Table~\ref{tab:tsm-languages} (\texttt{tsm}), and Table~\ref{tab:universal-ner-languages} (\texttt{universal-ner}).

\begin{table}[htbp]
\centering
\small
\begin{tabular}{@{}llllll@{}}
\toprule
\textbf{Code} & \textbf{Language} & \textbf{Code} & \textbf{Language} & \textbf{Code} & \textbf{Language} \\
\midrule
af-ZA & Afrikaans & it-IT & Italian & ru-RU & Russian \\
am-ET & Amharic & ja-JP & Japanese & sl-SL & Slovanian \\
ar-SA & Arabic & jv-ID & Javanese & sq-AL & Albanian \\
az-AZ & Azeri & ka-GE & Georgian & sv-SE & Swedish \\
bn-BD & Bengali & km-KH & Khmer & sw-KE & Swahili \\
ca-ES & Catalan & ko-KR & Korean & hi-IN & Hindi \\
zh-CN & Chinese (China) & lv-LV & Latvian & kn-IN & Kannada \\
zh-TW & Chinese (Taiwan) & mn-MN & Mongolian & ml-IN & Malayalam \\
da-DK & Danish & ms-MY & Malay & ta-IN & Tamil \\
de-DE & German & my-MM & Burmese & te-IN & Telugu \\
el-GR & Greek & nb-NO & Norwegian & th-TH & Thai \\
en-US & English & nl-NL & Dutch & tl-PH & Tagalog \\
es-ES & Spanish & pl-PL & Polish & tr-TR & Turkish \\
fa-IR & Farsi & pt-PT & Portuguese & ur-PK & Urdu \\
fi-FI & Finnish & ro-RO & Romanian & vi-VN & Vietnamese \\
fr-FR & French &  he-IL & Hebrew  & cy-GB & Welsh \\
 hu-HU &  Hungarian & hy-AM & Armenian & id-ID & Indonesian \\
 is-IS & Icelandic &  &  &  & \\
\bottomrule
\end{tabular}
\caption{Language codes and corresponding language names in \texttt{massive} dataset.}
\label{tab:language-codes}
\end{table}

\end{document}